\documentclass[conference,twoside]{IEEEtran}
\IEEEoverridecommandlockouts

\usepackage{cite}
\usepackage{amsmath,amssymb,amsfonts}
\usepackage{algorithm}
\usepackage{algorithmic}
\usepackage{graphicx}
\usepackage{textcomp}
\usepackage{xcolor}
\usepackage{subcaption}
\usepackage{hyperref}
\usepackage{tikz}
\usepackage{fancyhdr}
\usetikzlibrary{arrows.meta, positioning}

\usepackage{times} 

\def\BibTeX{{\rm B\kern-.05em{\sc i\kern-.025em b}\kern-.08em
    T\kern-.1667em\lower.7ex\hbox{E}\kern-.125emX}}

\title{Optimizing Deep Neural Networks using\\
Safety-Guided Self Compression}

\author{
    \IEEEauthorblockN{Mohammad Zbeeb\textsuperscript{1}, Mariam Salman\textsuperscript{2}, Mohammad Bazzi\textsuperscript{3}, Ammar Mohanna\textsuperscript{4*}}
    \IEEEauthorblockA{
        \textsuperscript{1,2,3,4}Department of Electrical and Computer Engineering, American University of Beirut (AUB)\\
        Emails: \{mbz02, mcs12, mab101, am288\}@mail.aub.edu
    }
    \thanks{*Corresponding Author}
}

\begin{document}

\fancyhead{}                      
\fancyhead[L]{\rule{0.4pt}{1em}\hspace{0.6em}\small\itshape Preprint - American University of Beirut}
\maketitle

\begin{abstract}
The deployment of deep neural networks on resource-constrained devices necessitates effective model compression strategies that judiciously balance the reduction of model size with the preservation of performance. This study introduces a novel safety-driven quantization framework that leverages preservation sets to systematically prune and quantize neural network weights, thereby optimizing model complexity without compromising accuracy. The proposed methodology is rigorously evaluated on both a convolutional neural network (CNN) and an attention-based language model, demonstrating its applicability across diverse architectural paradigms. Experimental results reveal that our framework achieves up to a 2.5\% enhancement in test accuracy relative to the original unquantized models while maintaining 60\% of the initial model size. In comparison to conventional quantization techniques, our approach not only augments generalization by eliminating parameter noise and retaining essential weights but also reduces variance, thereby ensuring the retention of critical model features. These findings underscore the efficacy of safety-driven quantization as a robust and reliable strategy for the efficient optimization of deep learning models. The implementation and comprehensive experimental evaluations of our framework are publicly accessible at \href{https://github.com/Moe-Zbeeb/Optimizing-Deep-Neural-Networks-via-Safety-Guided-Self-Compression.git}{GitHub}.
\end{abstract}


\section{Introduction}
Deep learning models are often framed as the process of discovering the most efficient program that performs well on training data and generalizes effectively to unseen data \cite{10.1145/3551349.3556906}. This principle drives model compression, where the aim is to reduce network complexity while maintaining performance. Compression is critical for deploying deep networks on resource-constrained devices, such as smartphones, where both memory and computational power are limited. Beyond reducing size, compression can enhance generalization and preserve the model’s reliability and robustness---a concept we term safety-driven compression.

Traditional methods like pruning and quantization reduce model size and improve inference speed \cite{tang2024surveytransformercompression}. However, these techniques often compromise reliability by removing components crucial for generalization \cite{10378737}\cite{10.1145/3551349.3556906}, posing risks to model stability in real-world applications.

Early approaches to pruning, such as Optimal Brain Damage \cite{NIPS1989_6c9882bb}, introduced sparsity by removing unnecessary weights. Recent advances like channel pruning \cite{He_2017_ICCV} improve efficiency by eliminating redundant filters, but may also inadvertently remove essential features. Similarly, quantization-aware training (QAT) \cite{bengio2013estimating} allows gradient flow through low-precision weights, balancing performance and efficiency. While adaptive quantization techniques \cite{defossez2022differentiable} adjust bit depths dynamically, they face challenges at very low precision. Hybrid methods, like XNOR-Net \cite{Rastegari2016XNORNetIC}, combine pruning and quantization for hardware efficiency but often require specialized infrastructure.

To address these limitations, we propose a safety-driven self-compression approach. Unlike traditional methods, our approach adaptively prunes non-essential weights while preserving critical components by continuously evaluating performance on a carefully selected subset of data, which we call the ``preservation set.'' This allows the network to ``compress itself'' while maintaining performance, reducing variance between training and test results, and ensuring robustness during real-world deployment.

We applied this method to a convolutional neural network (CNN) trained on the MNIST \cite{deng2012mnist} dataset and a decoder-based attention model. In both cases, pruning selectively targeted CNN kernels and Transformer’s attention heads while preserving critical features. Our results show that safety-driven compression achieves efficient and reliable optimization without sacrificing accuracy, eventually optimizing for accuracy.

\section{Methodology}

Traditional compression methods often risk the unintended removal of essential weights, potentially degrading model performance. To mitigate this, we propose a safety-driven self-compression approach that leverages a ``preservation set''---a strategically selected subset of data that safeguards critical network components throughout the compression process. By evaluating the model’s performance on this preservation set at each pruning step, our approach selectively eliminates only redundant parameters, ensuring that noise is removed without compromising the integrity of essential features.

In both vision and language models, the preservation set is developed using Grad-CAM \cite{Selvaraju_2019}, uncertainty sampling \cite{liu2023understandinguncertaintysampling}, and clustering techniques to identify representative data points. Despite differing modalities, both model types share a numerical data structure: in vision models, inputs are structured as grids representing pixel intensities, whereas in language models, inputs are encoded as numerical vectors within an embedding space. This shared structural framework facilitates the application of the preservation set methodology across both model types.

Grad-CAM (Gradient-weighted Class Activation Mapping) is employed to identify high-activation regions within image grids that are essential for accurate model predictions, thereby ensuring that the preservation set includes key visual features. In language models, Grad-CAM similarly highlights tokens or phrases that substantially influence the model’s output. To enhance the robustness of the preservation set, uncertainty sampling is applied to capture data points where the model exhibits the highest uncertainty, effectively addressing challenging cases prone to errors. Finally, clustering techniques promote diversity within the preservation set by grouping semantically distinct patterns, ensuring a comprehensive representation of the dataset’s intrinsic features.

Our method integrates these techniques to construct a preservation set that is both robust and representative, establishing a dependable standard for assessing the impact of quantization and pruning. Notably, the preservation set is formulated independently of the quantization process, allowing it to be flexibly applied across both vision and language models.
\begin{figure}[htb]
    \centering
    \begin{subfigure}[t]{0.48\textwidth}  
        \includegraphics[width=\textwidth, height=0.18\textheight]{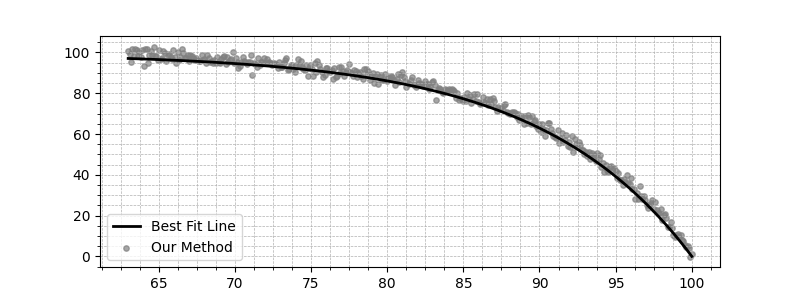}  
        \caption{Effect of pruning on training performance.}
        \label{fig:pruning}
    \end{subfigure}
    \hfill
    \begin{subfigure}[t]{0.48\textwidth}  
        \includegraphics[width=\textwidth, height=0.18\textheight]{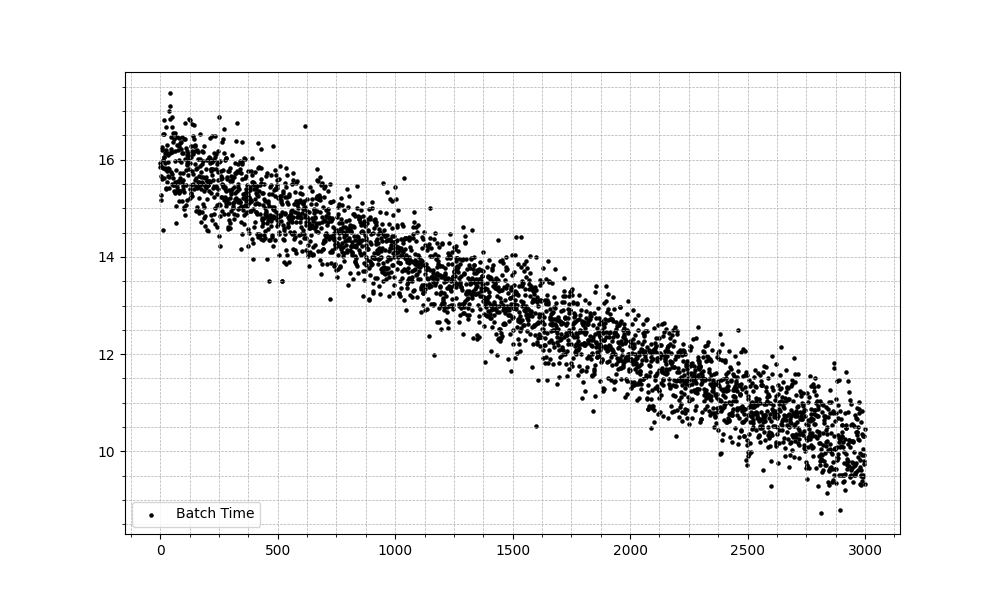}  
        \caption{Acceleration of training time as parameters are pruned.}
        \label{fig:training_time}
    \end{subfigure}
    \caption{(a) The solid black line represents the best fit to the data, showcasing how the model retains its predictive capacity, while the gray dots represent results obtained using our proposed method. As the parameters are pruned, the network's size is reduced rapidly during the early stages of training, followed by further compression. Importantly, this approach preserves test performance, maintaining accuracy while preventing any significant increase in variance. (b) The plot illustrates how training time accelerates as parameters are pruned from the network. As the number of iterations increases, the network becomes more efficient, leading to a significant reduction in training time (sampled 3000 data points).}
    \label{fig:side_by_side}
\end{figure}
\subsection*{Differentiable Quantization}
The core of our compression method is a differentiable quantization function \cite{cséfalvay2023selfcompressingneuralnetworks}, designed to dynamically adjust the bit depth of the neural network’s weights throughout training. This function is defined as:
\begin{equation}
q(x, b, e) = 2^e \left\lfloor \min \left( \max \left( 2^{-e} x, -2^{b-1} \right), 2^{b-1} - 1 \right) \right\rfloor, 
\end{equation}
where \( x \) represents the input values (weights), \( b \) denotes the bit depth, and \( e \) serves as the scaling exponent. Through the application of a \( 2^e \) scaling factor, this function compresses the network’s parameters while clamping values within a defined range, effectively mapping input values into a discrete set constrained by \( -2^{b-1} \) and \( 2^{b-1} - 1 \).

The differentiability of this quantization function is facilitated by the Straight-Through Estimator (STE) \cite{yin2019understandingstraightthroughestimatortraining}, an approximation method that overcomes the inherent non-differentiability of the rounding operation in quantization. By employing STE, gradients are able to propagate through the quantization step during backpropagation, thereby enabling a dynamic adjustment of bit depth within the gradient-based learning process. This approach integrates quantization directly into the training phase, allowing the network to adaptively modify its bit depth in response to variations in data and loss function, enhancing both the flexibility and efficiency of the compression process.

The overall optimization objective extends the standard loss function by introducing additional terms that simultaneously encourage compression and preserve critical features identified by the preservation set. The total loss function is defined as:
\begin{equation}
\Lambda(x) = \sum_{i=1}^{N} \ell(y_i, \hat{y}_i) + \alpha \sum_{l=1}^{L} ||W^l||_1 + \gamma Q + \lambda \cdot \text{preservation loss}
\end{equation}
where \( \ell(y_i, \hat{y}_i) \) represents the primary prediction loss, such as cross-entropy or mean squared error, depending on the task at hand. The term \( \alpha \sum_{l=1}^{L} ||W^l||_1 \) is an L1 regularization term promoting sparsity by penalizing the sum of absolute weights across layers, which aids in model simplification. The component \( \gamma Q \) penalizes excessive model size by discouraging high bit depths, thereby incentivizing quantization. Finally, \( \lambda \cdot \text{preservation loss} \) ensures that the model retains performance on the preservation set, thereby safeguarding essential features during the compression process.

This novel formulation frames the optimization problem in three distinct but interdependent goals: first, ensuring strong performance on the primary learning task; second, optimizing the size of the network by balancing sparsity and quantization; and third, maintaining high performance on the preservation set, thus reinforcing the retention of critical model capacities.

What distinguishes this approach is its multi-faceted balancing act: the model is not only encouraged to achieve accuracy on the primary task, but is also rewarded for compressing parameters without compromising core capabilities. By targeting these three objectives within a single loss function, our method uniquely addresses the challenge of maintaining model reliability in compressed formats, providing a more robust solution than traditional approaches that optimize solely for accuracy or compression alone. This design not only results in a smaller model but also ensures that preserved features enable the model to generalize effectively, making this framework particularly suitable for applications where efficiency and reliability are paramount.

The average bit depth \( \boldsymbol{Q} \) \cite{7900188} is introduced as a metric to assess the mean quantization level across all layers, providing an overall measure of the model’s compression. Specifically, \( N \) denotes the total number of layers, while \( \boldsymbol{z}_l \) represents the quantized size of each individual layer \( l \), allowing us to express \( \boldsymbol{Q} \) as:
\begin{equation}
\boldsymbol{Q} = \frac{1}{N} \sum_{l=1}^{L} \boldsymbol{z}_l,
\end{equation}
where \( \boldsymbol{z}_l \), the quantized size of layer \( l \) within the vision model, is calculated as:
\begin{equation}
\boldsymbol{z}_l = \boldsymbol{I}_l \boldsymbol{H}_l \boldsymbol{W}_l \sum_{i=1}^{O_l} \boldsymbol{b}_{l,i}.
\end{equation}  

In this context:
\begin{itemize}
    \item \( \boldsymbol{I}_l \) denotes the number of input channels for layer \( l \),
    \item \( \boldsymbol{H}_l \) and \( \boldsymbol{W}_l \) refer to the spatial dimensions of the layer’s output, specifically its height and width, which define the output feature map’s resolution,
    \item \( \boldsymbol{O}_l \) corresponds to the number of output channels, indicating the layer’s capacity for feature extraction, and
    \item \( \boldsymbol{b}_{l,i} \) specifies the bit depth assigned to the \( i \)-th output channel, thus allowing for channel-specific precision adjustments across layers.
\end{itemize}

This formulation provides a comprehensive means of adjusting the quantization at each layer, accommodating varying bit depths within individual channels. By dynamically adjusting \( \boldsymbol{b}_{l,i} \), our model achieves a fine-grained control over compression, optimizing the bit depth based on the preservation and performance requirements at each layer.

For the decoder model, the model size is dictated by the configuration of its attention layers, which includes key parameters such as the number of attention heads, the dimension of each head, the size of the hidden layers, and the overall number of layers. Each attention layer utilizes multiple heads to process input sequences, allowing the model to concurrently attend to different aspects of the input data. Consequently, the model’s size is substantially influenced by the cumulative dimensions of these attention mechanisms, along with the subsequent feedforward layers.

Applying quantization to these components achieves a compact representation of the transformer model, retaining its capability to process and interpret complex sequences while reducing memory requirements. Notably, compressing the feedforward layers, although beneficial, would necessitate specialized hardware, which is beyond the scope of this experiment. Therefore, in the current setup, we focus on quantizing the attention layers, allowing for effective compression without imposing additional hardware constraints.

The core concept of our approach is that the parameters guiding model quantization are entirely learnable and can be integrated into the training process via a differentiable mechanism. By allowing the quantization parameters to be dynamically learned during training, the model can adapt its precision requirements based on data-driven insights, thereby optimizing for a balance between model size reduction and performance retention. This approach is operationalized through a carefully designed loss function that incorporates multiple objectives: the primary prediction loss, a sparsity-inducing penalty, and a preservation set loss.

The preservation set, acting as a regularization component, provides feedback that directs the compression process. Specifically, it monitors the effects of quantization on critical network functions, ensuring that only redundant components are pruned while vital structures are retained. By integrating the preservation set loss into the total loss function, we create a mechanism that prevents the compression process from compromising model reliability and accuracy. This safeguard, represented mathematically within the loss function, allows the model to selectively adjust bit precision across layers or specific components, such as kernels in convolutional layers or attention heads in transformers.

The quantization step itself is controlled by an additional learnable parameter, which dynamically determines the number of bits required to encode each kernel or attention head. This parameter is adjusted iteratively throughout training based on feedback from the preservation set. By continuously assessing the preservation set’s performance, the model can recalibrate the quantization levels, restoring higher precision when necessary to maintain critical features. Consequently, our approach promotes compression that is both efficient and adaptable to the specific requirements of the data, achieving an optimized balance between model size and accuracy.

To realize this framework, we implement an enhanced training loop. This loop systematically applies our proposed loss function, evaluates quantization effects, and adjusts model parameters accordingly. At each step, it performs a compression step that prunes redundant weights and removes zeroed-out components (such as kernels or attention heads). If the preservation set’s performance degrades, precision is restored to affected components, ensuring that essential features are maintained. This iterative loop promotes an adaptive compression process, yielding a compact yet high-performing model.

\begin{algorithm}
\caption{Safety-Driven Self-Compression Training Loop}
\begin{algorithmic}[1]
\STATE Initialize model weights and bit precision settings $\{b_i\}_{i=1}^L$ for each layer $i$
\FOR{each training epoch}
    \FOR{each mini-batch of data $(X, Y)$}
        \STATE Forward pass through the model with current bit precision $\{b_i\}$
        \STATE Compute the prediction loss $\ell(Y, \hat{Y})$
        \STATE Compute L1 regularization term $\alpha \sum_{l=1}^L \|W^l\|_1$
        \STATE Compute quantization penalty $\gamma Q$ based on current bit depths
        \STATE Compute preservation loss $\lambda \cdot \text{preservation loss}$ using the preservation set
        \STATE Total loss $\Lambda = \ell(Y, \hat{Y}) + \alpha \sum_{l=1}^L \|W^l\|_1 + \gamma Q + \lambda \cdot \text{preservation loss}$
        
        \STATE Backpropagate the total loss $\Lambda$ and update weights and bit precision parameters $\{b_i\}$
        
        \STATE Apply quantization: adjust weights to match target bit precision $\{b_i\}$
        \FOR{each layer $l$}
            \IF{all weights in kernel/attention head are zero}
                \STATE Prune kernel/attention head from layer $l$
            \ENDIF
        \ENDFOR
        
        \IF{performance on the preservation set drops below threshold}
            \STATE Increase bit precision $\{b_i\}$ for affected components in layer $l$
        \ENDIF
    \ENDFOR
\ENDFOR
\end{algorithmic}
\end{algorithm}

\section{Experimental Setup}

In this study, we assess a preservation-driven quantization approach across two distinct neural network architectures: a compact Convolutional Neural Network (CNN) trained on the MNIST \cite{deng2012mnist} dataset and a Transformer model trained on a dataset of names for n-gram analysis, a standard benchmark in language modeling tasks. The primary objective is to achieve an optimal balance between network compression and the retention of critical components, thereby ensuring stable and reliable performance despite the application of aggressive pruning and quantization.

The CNN architecture comprises \( n \) quantized convolutional layers — a hyperparameter fine-tuned and averaged to achieve more stable and robust results. Each convolutional layer employs ReLU activations, followed by batch normalization to stabilize and accelerate the training process, with max pooling layers to reduce spatial dimensions. These layers are connected to a fully connected layer for classification, creating a straightforward yet effective architecture. This design is particularly suited to evaluating the impact of quantization on visual data processing, providing an efficient framework for testing the efficacy of compression techniques in maintaining model performance.

The Transformer model leverages self-attention mechanisms to process tokenized text, establishing a robust and scalable framework for language modeling tasks. Central to this architecture are several key components: an embedding layer that encodes token and positional information, a series of Transformer blocks each composed of multi-head self-attention and feed-forward layers with integrated layer normalization, and a final linear layer responsible for output generation. The model’s reliance on self-attention enables it to effectively manage and capture long-range dependencies within textual data, making it particularly suitable for quantization studies aimed at preserving accuracy in complex sequence processing.

A central component of our experimental methodology is the integration of preservation sets for both model architectures, designed to rigorously evaluate the resilience of quantization processes. For the CNN, the preservation set includes 10\% of the training images, systematically selected based on criteria defined in prior sections, ensuring the representation of critical visual features. In the Transformer model, the preservation set is constructed from carefully chosen sentences exhibiting essential linguistic characteristics, thus safeguarding the integrity of key syntactic and semantic structures. These preservation sets are instrumental in maintaining the models’ functional integrity, as they provide benchmarks that prevent the loss of crucial model behavior throughout the quantization process.
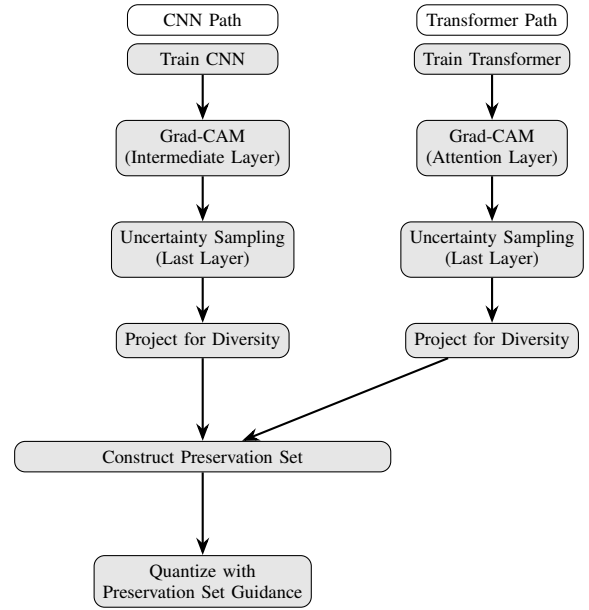
\begin{figure}[ht]
\centering
\begin{tikzpicture}[
    node distance=0.6cm and 0.8cm,
    every node/.style={rectangle, rounded corners, draw, minimum width=2cm, align=center, font=\scriptsize},
    arrow/.style={-{Stealth}, thick},
    process/.style={fill=gray!20}
]

\node (cnn_start) [process] {Train CNN};
\node (cnn_gradcam) [process, below=of cnn_start] {Grad-CAM \\ (Intermediate Layer)};
\node (cnn_uncertainty) [process, below=of cnn_gradcam] {Uncertainty Sampling \\ (Last Layer)};
\node (cnn_project) [process, below=of cnn_uncertainty] {Project for Diversity};

\node (trans_start) [process, right=of cnn_start, xshift=1cm] {Train Transformer};
\node (trans_gradcam) [process, below=of trans_start] {Grad-CAM \\ (Attention Layer)};
\node (trans_uncertainty) [process, below=of trans_gradcam] {Uncertainty Sampling \\ (Last Layer)};
\node (trans_project) [process, below=of trans_uncertainty] {Project for Diversity};

\node (combine) [process, below=of cnn_project, yshift=-0.5cm, minimum width=5cm] {Construct Preservation Set};
\node (quantize) [process, below=of combine, yshift=-0.5cm] {Quantize with \\ Preservation Set Guidance};

\draw [arrow] (cnn_start) -- (cnn_gradcam);
\draw [arrow] (cnn_gradcam) -- (cnn_uncertainty);
\draw [arrow] (cnn_uncertainty) -- (cnn_project);

\draw [arrow] (trans_start) -- (trans_gradcam);
\draw [arrow] (trans_gradcam) -- (trans_uncertainty);
\draw [arrow] (trans_uncertainty) -- (trans_project);

\draw [arrow] (cnn_project) -- (combine);
\draw [arrow] (trans_project) -- (combine);
\draw [arrow] (combine) -- (quantize);

\node [above=0.1cm of cnn_start, font=\scriptsize] {CNN Path};
\node [above=0.1cm of trans_start, font=\scriptsize] {Transformer Path};

\end{tikzpicture}
\caption{Flowchart depicting the creation of the preservation set for CNN and Transformer models: Grad-CAM, uncertainty sampling, and projection at the last layer ensure diversity, with the preservation set guiding quantization.}
\label{fig:preservation_set_quantization}
\end{figure}
The experimental framework involved iterative trials, where evaluations were conducted at predefined intervals to systematically monitor critical performance metrics. The training process employed an optimized composite loss function, which seamlessly integrates multiple objectives: a prediction loss component to measure the discrepancy between model predictions and ground truth labels; a sparsity-inducing regularization term aimed at promoting model sparsity; a quantization penalty that encourages lower bit representations to facilitate compactness; and a preservation loss, specifically incorporated to ensure that model performance on the preservation set remains uncompromised. This multifaceted loss function fosters the development of compact yet robust models, effectively balancing the goals of efficiency and reliability.

Quantization was applied selectively to the convolutional layers in the CNN and the attention heads in the Transformer, utilizing pruning based on feedback from the preservation sets to protect critical model components. A dynamic feedback mechanism was implemented to monitor accuracy on the preservation sets. If accuracy fell below a predefined threshold, pruned weights were restored to their original precision, thereby mitigating the potential risks associated with excessive quantization and ensuring the preservation of essential model functionalities.

To further enhance robustness and generalizability, the experiments were conducted across a broad range of hyperparameter settings. Batch sizes of 16, 32, 64, and 128 were tested to evaluate the influence of processing scale on quantization performance. Learning rates between \(1 \times 10^{-4}\) and \(1 \times 10^{-2}\) were explored to optimize training dynamics under quantization. Furthermore, bit depths of 2, 4, 8, and 16 bits were analyzed to assess the effects of various compression levels on model performance, while layer numbers were varied to study the resilience of quantization in different model depths. Experiments were also conducted on both CPU and GPU environments to ensure consistent performance across hardware platforms. Results were averaged across these configurations to provide a holistic view of the method's efficacy.

The key performance metrics monitored included test accuracy, model size (measured in bit depth and memory footprint), and preservation loss, which quantifies the impact of quantization on the preservation sets. These metrics offered nuanced insights into the trade-offs between compression and performance retention.

Particular emphasis was placed on the Transformer model due to its reliance on attention mechanisms. Quantization and pruning were applied specifically to the attention heads to rigorously assess the model’s capacity to maintain long-range dependencies in text under quantization. Comparative analysis between GPU and CPU runs provided consistency across hardware platforms.

Throughout the training process, continuous monitoring of these metrics ensured that core model capabilities were preserved despite aggressive compression. By averaging results across a diverse array of hyperparameter settings and hardware environments, we ensured that the observed effects of quantization remained reliable and independent of specific configurations. The resulting compressed models achieved an optimal balance between size reduction and predictive accuracy, affirming the robustness and versatility of the preservation-driven quantization approach across different architectures and platforms.

\section{Results}

Neural networks are powerful tools that inherently balance model size with predictive performance by capturing complex data relationships. This balance is particularly crucial in the context of quantization and pruning techniques, where achieving optimal trade-offs between model complexity and accuracy is essential for reliable deployment in resource-constrained environments.

Our experimental results are averaged across a comprehensive set of hyperparameter settings, including variations in layer depth, learning rates, and bit depths, to ensure a robust evaluation of each quantization approach across vision and language models.

On average, the safety-driven quantization models retain approximately 60\% of the original model size while outperforming both the unquantized and unsafely quantized models in test performance. Specifically, the Safety-Driven Quantization method reduced the model size from 326,192 bytes to 214,730 bytes for the convolutional-based vision model and from 831,846 bytes to 505,381 bytes for the attention-based decoder. Notably, test accuracy for the vision model improved from 98.6\% to 99.5\%, while test loss for the decoder model decreased from 1.8 to 1.6. These improvements indicate that our proposed approach successfully reduces model complexity while enhancing predictive accuracy.

The significance of these results lies in the surpassing of the original, unquantized model's performance, which implies not only effective compression but also improved generalization and reduced variance. The safety-driven quantization framework retains critical model functionalities by focusing on essential features, which enables the model to generalize more effectively. This reduction in variance further underscores the stability of the compressed models, ensuring reliable deployment without sacrificing accuracy.

Our findings underscore the efficacy of intelligent, preservation-focused pruning strategies during training. Smaller, well-optimized models can indeed surpass larger, uncompressed counterparts in performance, as evidenced by our approach. This strategy presents a transformative method for model compression, achieving substantial size reductions efficiently while preserving, or even enhancing, performance.

\begin{table}[htbp]
\centering
\renewcommand{\arraystretch}{1.2}  
\scriptsize  
\begin{tabular}{|l|c|c|}
\hline
\textbf{Model Type} & \textbf{Test Accuracy (\%)} & \textbf{Model Size (bytes)} \\ \hline
Without Quantization & 98.6 & 326,192 \\ \hline
Unsafe Quantization \cite{cséfalvay2023selfcompressingneuralnetworks} & 97.0 & 210,404 \\ \hline
Safety-Driven Quantization & 99.5 & 214,730 \\ \hline
\end{tabular}
\caption{Comparison of model performance and size across quantization approaches for the Convolutional-Based Vision Model.}
\label{tab:vision_results}
\end{table}

\begin{table}[htbp]
\centering
\renewcommand{\arraystretch}{1.2}  
\scriptsize  
\begin{tabular}{|l|c|c|}
\hline
\textbf{Model Type} & \textbf{Test Loss} & \textbf{Model Size (bytes)} \\ \hline
Without Quantization & 1.8 & 831,846 \\ \hline
Unsafe Quantization \cite{cséfalvay2023selfcompressingneuralnetworks} & 1.9 & 425,865 \\ \hline
Safety-Driven Quantization & 1.6 & 505,381 \\ \hline
\end{tabular}
\caption{Comparison of model performance and size across quantization approaches for the Attention-Based Decoder.}
\label{tab:decoder_results}
\end{table}

\noindent As shown in \textbf{Table~\ref{tab:vision_results}} for the Convolutional-Based Vision Model and in \textbf{Table~\ref{tab:decoder_results}} for the Attention-Based Decoder, the safety-driven quantization approach not only achieves significant model size reductions but also maintains or improves upon the performance of the original models. These results illustrate that safety-driven quantization is a robust and effective strategy for preserving essential features, enhancing generalization, and reducing variance across diverse model architectures.
\begin{figure}[ht]
    \centering
    \includegraphics[width=0.8\linewidth]{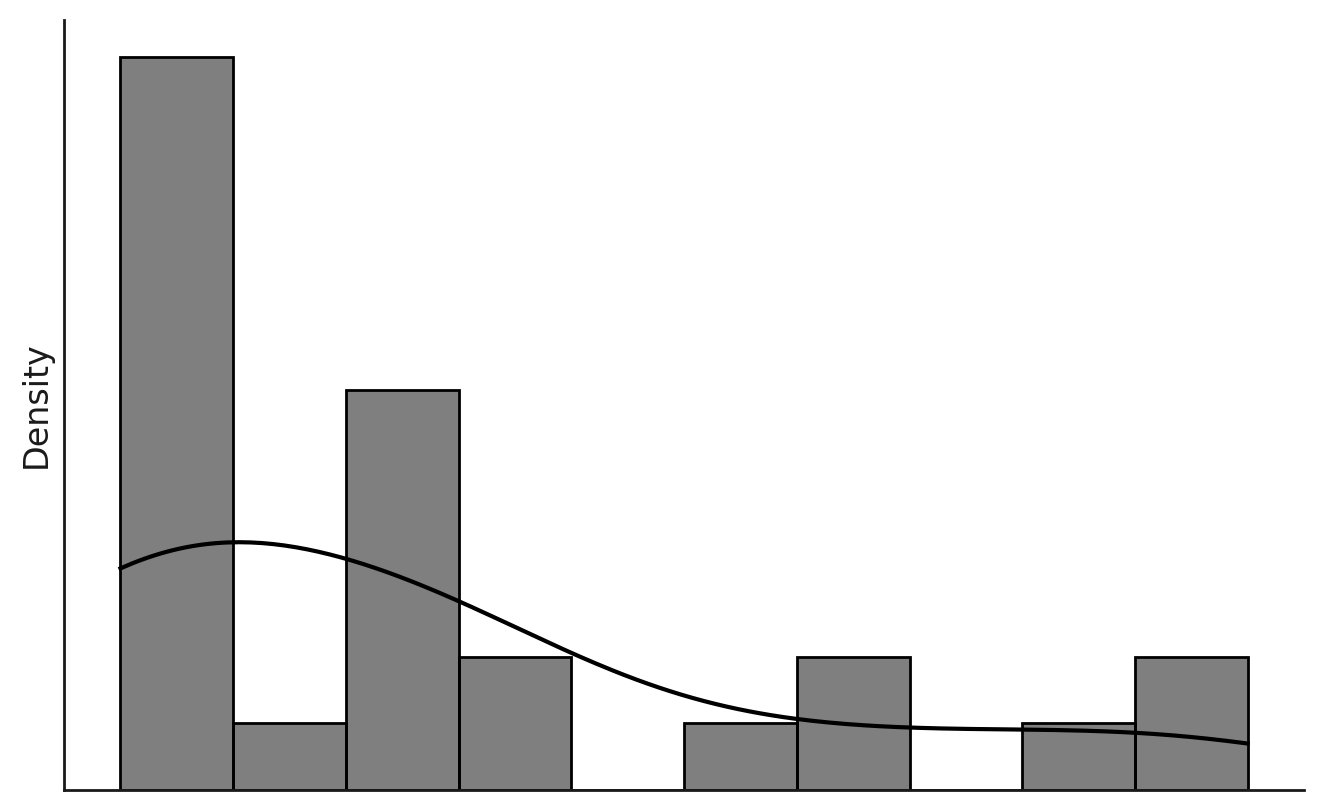}
\caption{Histogram with density curve of neural network weights post-quantization. The distribution shows a concentration of weights near central values, indicating effective pruning and quantization of less impactful weights. The tails represent significant weights preserved to maintain essential model features, supporting generalization and reducing variance. This distribution underscores the effectiveness of safety-driven quantization in achieving efficient model compression without sacrificing accuracy.}

    \label{fig:weights_distribution}
\end{figure}

\section{Conclusion}

The results of this study conclusively demonstrate the value of the proposed safety-driven quantization framework, showcasing its effectiveness in reducing performance variance between training and testing phases and thereby promoting both stronger generalization and minimized overfitting. By selectively pruning less significant weights, this approach manages to retain critical model accuracy while efficiently reducing model complexity, presenting a solution well-suited to deployment on resource-constrained platforms without sacrificing performance.

This research highlights the importance of quantization, not only as a tool for model size reduction but as a crucial methodology for enhancing the reliability and efficacy of deep learning architectures. The safety-driven approach adds a novel dimension to conventional compression methods by integrating a preservation set, which carefully monitors and safeguards core model features. This ensures that essential components remain intact and that the model's robustness and interpretability are preserved under compression—a critical advancement over standard techniques, which often lack this precision.

The findings emphasize the potential of safety-driven quantization as an effective, adaptable strategy for addressing modern machine learning challenges. Future research will extend this framework to more complex neural architectures and real-world application contexts, further examining its ability to optimize model performance and reliability under diverse conditions. The balance achieved between compactness and performance retention demonstrates this approach’s suitability for robust and efficient neural network deployment across a range of real-world, resource-limited environments, potentially setting a new standard for model optimization in practical AI applications.  

Limitations of the proposed quantization approach is that it requires specialized hardware for application to feed-forward dense layers, and manual adjustments to the training loop, as it is not yet supported in standard deep learning frameworks.

\bibliographystyle{IEEEtran}
\bibliography{main}

\end{document}